\pdfoutput=1
\documentclass[11pt]{article}

\usepackage{emnlp2021}

\usepackage{times}
\usepackage{latexsym}

\usepackage[parfill]{parskip}
\usepackage[utf8]{inputenc} % allow utf-8 input
\usepackage[T1]{fontenc}    % use 8-bit T1 fonts
\usepackage{hyperref}       % hyperlinks
\usepackage{url}            % simple URL typesetting
\usepackage{booktabs}       % professional-quality tables
\usepackage{amsfonts}       % blackboard math symbols
\usepackage{nicefrac}       % compact symbols for 1/2, etc.
\usepackage{microtype}      % microtypography
\usepackage{graphicx}
\usepackage{natbib}
\usepackage{tabularx}
\usepackage{xtab}
\usepackage{todonotes}
\usepackage{amsmath}

\usepackage[T1]{fontenc}
\usepackage[utf8]{inputenc}

\title{An Empirical Exploration in Quality Filtering of Text Data}

\author{Leo Gao \\
  EleutherAI \\
  \texttt{lg@eleuther.ai}
}

\begin{document}
\maketitle
\begin{abstract}
While conventional wisdom suggests that more aggressively filtering data from low-quality sources like Common Crawl always monotonically improves the quality of training data, we find that aggressive filtering can in fact lead to a decrease in model quality on a wide array of downstream tasks for a GPT-like language model. We speculate that this is because optimizing sufficiently strongly for a proxy metric harms performance on the true objective, suggesting a need for more robust filtering objectives when attempting to filter more aggressively. We hope this work leads to detailed analysis of the effects of dataset filtering design choices on downstream model performance in future work.
\end{abstract}

\section{Introduction}

As language models increase in size, the need for large, high-quality text datasets has increased as well. Recent work in dataset construction for large language models has centered largely on taking large internet corpora like Common Crawl and employing some method of filtering using some proxy for quality to extract a smaller, high quality training set \citep{Wenzek2019CCNet__Extracting_High_Quality,Brown2020Language_Models_are_Few_Shot_L,Raffel2020Exploring_the_Limits_of_Transf,Yang2020XLNet__Generalized_Autoregress}. In particular, we focus on shallow classifier-based quality filtering as in \citet{Brown2020Language_Models_are_Few_Shot_L} because it provides a simple, continuous, and quantifiable way to adjust the aggressiveness of filtering, and because this reflects the type of classifier used in prior work.

While intuitively it may seem like the more data is discarded the higher quality the remaining data will be, we find that this is not always the case with shallow classifier-based filtering. Instead, we find that filtering improves downstream task performance up to a point, but then decreases performance again as the filtering becomes too aggressive.

We speculate that this decrease in performance is due to Goodhart's law \citep{Goodhart1984}, and specifically regressional Goodharting \citep{Manheim2019Categorizing_Variants_of_Goodh}:

\textbf{Goodhart's Law.} Any observed statistical regularity will tend to collapse once pressure is placed upon it for control purposes. \citep{Goodhart1984}

In other words, optimizing a metric that is a proxy for a desired outcome tends to invalidate the proxy. By optimizing too strongly for the classifier's score by discarding too many low-scoring documents, the documents that are kept are consistently biased towards the ones with features superficially resembling the high quality data in a way that satisfies the classifier, rather than truly high quality data. 

%\todo[inline]{The current plan: we're gonna train a 1B+ quality model with active learning from human feedback and hopefully show that it does way better at data filtering, and then we're going to put all of Pile and hopefully also CC through. That shifts the focus of the paper, andso most of what's currently here will actually be the section that proves why using such a sophisticated model is necessary}

\begin{figure}[t]
    \centering
    \includegraphics[width=0.5\textwidth]{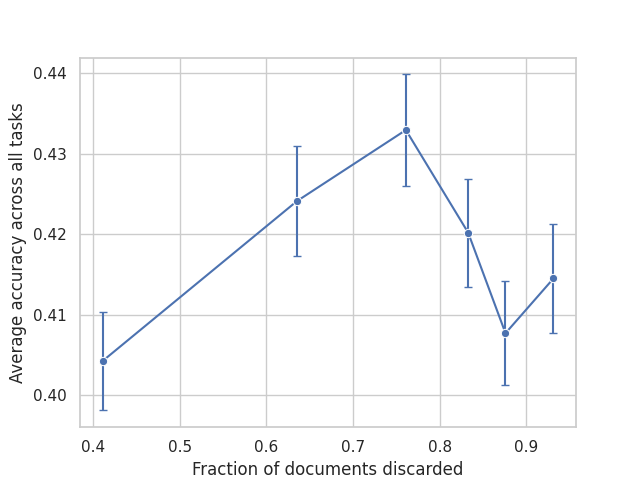}
    \caption{Average accuracy across all 13 tasks for various different filtering ratios using a shallow quality classifier.\footnotemark The amount of data post-filtering is held constant. Although filtering improves performance at first, discarding more data can actually reduce accuracy, due to misalignment between filtering classifier objective and text quality.}
    \label{fig:fig_filterratio_average.acc}
\end{figure}
\footnotetext{The average is taken across all task accuracies, with each task weighted equally. The error bars in this plot represent standard error and are computed by $\mathrm{se_{mean}} = n^{-1}\sqrt{\sum \mathrm{se}_i^2}$, where $\mathrm{se}_i$ represents the standard error for each individual task.}

\section{Related work}

The recent proliferation of ever larger language models has led to increasing demands on training data \citep{GPT,GPT2,OpenWebText,TuringNLG,Megatron,BERT,RoBERTa,Raffel2020Exploring_the_Limits_of_Transf,Brown2020Language_Models_are_Few_Shot_L,Zeng2021PanGu___Large_scale_Autoreg}. This data is increasingly derived from internet corpora like Common Crawl \citep{GPT2,fast-text,wenzek-etal-2020-ccnet,conneau-etal-2020-unsupervised,Brown2020Language_Models_are_Few_Shot_L,Gao2020The_Pile__An_800GB_Dataset_of_,Raffel2020Exploring_the_Limits_of_Transf}.

However, the quality of raw Common Crawl data is often insufficient to be directly used. To combat this, many existing works use some kind of proxy for quality, like a classifier between known high quality data and low quality data \citep{Brown2020Language_Models_are_Few_Shot_L, Gao2020The_Pile__An_800GB_Dataset_of_,Zeng2021PanGu___Large_scale_Autoreg}, hand-crafted heuristics \citep{Yang2020XLNet__Generalized_Autoregress,Raffel2020Exploring_the_Limits_of_Transf}, or keeping only documents with perplexity scores that fall in some middle quantile of an existing language model \citep{wenzek-etal-2020-ccnet}. \citet{Brown2020Language_Models_are_Few_Shot_L} in particular filter extremely aggressively using their classifier, discarding about 98.7\% of their data.

Previous work has shown that models trained on heuristic-filtered datasets perform better on downstream tasks \citep{Raffel2020Exploring_the_Limits_of_Transf}. However, \citet{Gao2020The_Pile__An_800GB_Dataset_of_} show that a perplexity-filtered CC-derived dataset actually performs worse than unfiltered CC on certain tasks. \citet{Brown2020Language_Models_are_Few_Shot_L} do not provide any detailed analysis, but claim better quality for filtered data as evaluated through loss on held out sets of ``generative text samples."

% \begin{figure}[t]
%     \centering
%     \includegraphics[width=0.5\textwidth]{figures/fig_filterratio_lambada.acc.png}
%     \caption{LAMBADA accuracy for various different filtering ratios. Performance drops sharply as more documents are discarded.}
%     \label{fig:fig_filterratio_lambada.acc}
% \end{figure}

% \begin{figure}[t]
%     \centering
%     \includegraphics[width=0.5\textwidth]{figures/fig_filterratio_pubmedqa.acc.png}
%     \caption{PubmedQA accuracy for various different filtering ratios. Performance drops sharply as more documents are discarded.}
%     \label{fig:fig_filterratio_pubmedqa.acc}
% \end{figure}

\section{Downstream Evaluation Experiment}

\begin{figure*}
    \centering
    \includegraphics[width=\textwidth]{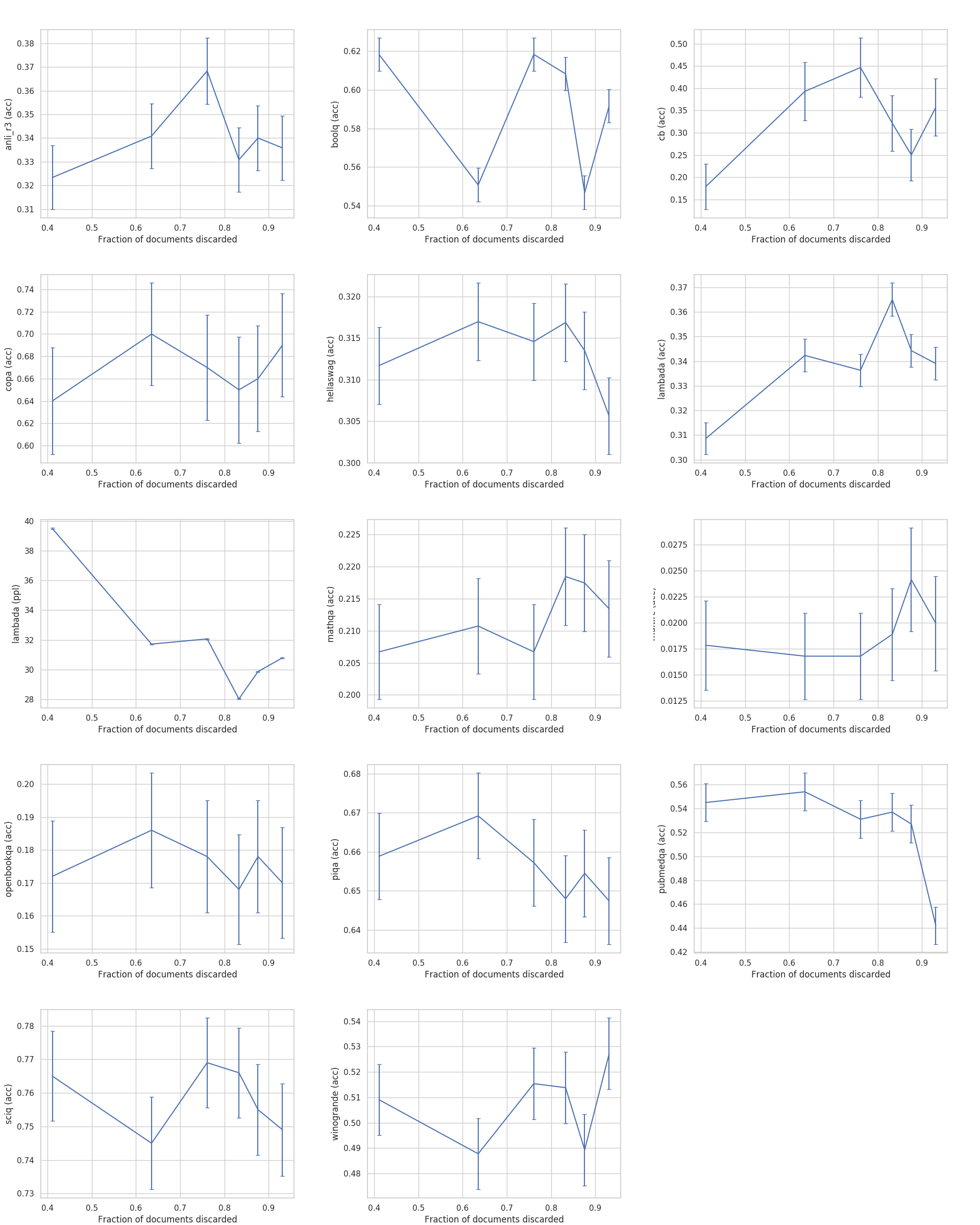}
    \caption{Plots of results for all downstream tasks explored in this paper. Higher is better on all metrics except LAMBADA perplexity (first plot in the third row), where lower is better.}
    \label{fig:montage}
\end{figure*}

To evaluate the effect of different degrees of filtering, we create a series of training sets with a controlled filtering methodology but with different hyperparameter settings to result in varied filtering ratios. We filter using the same method used in \citet{Brown2020Language_Models_are_Few_Shot_L}, with a Pareto-distribution thresholded filtering method and a shallow CommonCrawl-WebText classfier. In this method, rather than using a hard threshold, the threshold $\tau \sim \mathrm{Pareto}(\alpha)$ is sampled from a Pareto distribution, such that each document is kept if $\tau > 1 - \text{\texttt{score}}$, where $\alpha$ is a hyperparameter that controls the permissivity of the filter (see Table \ref{table:cc_alpha_ratio}). In effect, this relaxes the filter when compared to a hard threshold and allows some low-scoring data to be kept. 

\begin{table}
\centering
\begin{tabular}{c c} % centered columns (4 columns)
\toprule %inserts double horizontal lines
$\alpha$ & Fraction Discarded \\ [0.5ex] % inserts table
%heading
\midrule % inserts single horizontal line
1 & 0.4107 \\
2 & 0.6351 \\
3 & 0.7610 \\
4 & 0.8329 \\
5 & 0.8761 \\
6 & 0.9026 \\
7 & 0.9198 \\
8 & 0.9315 \\
%9 & 0.0602 \\
\bottomrule %inserts single line
\end{tabular}
\caption{Percentage of discarded documents of various settings using our classifier.} % title of Table
\label{table:cc_alpha_ratio} % is used to refer this table in the text
\end{table}

As none of the data or models used in \citet{Brown2020Language_Models_are_Few_Shot_L} has been made public, we instead use the same type of fasttext \citep{joulin2017bag} classifier between unfiltered Common Crawl and OpenWebText2 as used in \citet{Gao2020The_Pile__An_800GB_Dataset_of_}.

We use GPT-Neo \citep{gpt-neo} to train a series of models on each training set and evaluate on downstream tasks using the EleutherAI LM evaluation harness \citep{eval-harness}. Each model is 1.3 billion parameters, has a GPT-2 architecture \citep{GPT2} with the same model hyperparameters as the GPT-3-XL setting in \citet{Brown2020Language_Models_are_Few_Shot_L}, and is trained for 25k iterations with a batch size of 256.

% \begin{figure}
%     \centering
%     \includegraphics[width=0.5\textwidth]{figures/fig_filterratio_piqa.acc.png}
%     \caption{PiQA accuracy for various different filtering ratios. The optimal $\alpha$ is significantly different from LAMBADA or ANLI Round 3.}
%     \label{fig:fig_filterratio_piqa.acc}
% \end{figure}

To ensure that the effect is not confined to any specific task. we evaluate on a series of many downstream tasks. We use zero-shot prompting with no task-specific fine tuning and with prompting inspired by \citet{Brown2020Language_Models_are_Few_Shot_L} for many tasks.
In total, we evaluate on ANLI Round 3 \citep{Nie2020Adversarial_NLI__A_New_Benchma}, BoolQ \citep{clark-etal-2019-boolq}, CommitmentBank \citep{de_Marneffe_Simons_Tonhauser_2019}, COPA \citep{gordon-etal-2012-semeval}, Hellaswag \citep{zellers-etal-2019-hellaswag}, LAMBADA \citep{paperno-etal-2016-lambada}, MathQA \citep{amini-etal-2019-mathqa}, MultiRC \citep{khashabi-etal-2018-looking}, OpenbookQA \citep{mihaylov-etal-2018-suit}, PiQA \citep{Bisk2019PIQA__Reasoning_about_Physical}, PubmedQA \citep{Jin2019PubMedQA__A_Dataset_for_Biomed}, SciQ \citep{welbl-etal-2017-crowdsourcing}, and Winogrande \citep{Sakaguchi2019WinoGrande__An_Adversarial_Win}. Error bars in all evaluation task plots indicate standard error with respect to instances of the evaluation task.

For the training data, we create 40 GB filtered chunks of the Common Crawl data for each value of $\alpha \in \{1, 2, 3, 4, 5, 8\}$; in other words, different amounts of raw Common Crawl data are consumed for different $\alpha$ to produce the same fixed 40GB size result. For reference, \citet{Brown2020Language_Models_are_Few_Shot_L} filter even more aggressively than we do, discarding about 98.7\% of their data. The 40GB size is chosen because it is approximately the size of OpenWebText, which is representative of the amount of data usually used to train models of this size.
% https://arxiv.org/pdf/1910.14599.pdf
% https://www.aclweb.org/anthology/N19-1300/
% https://ojs.ub.uni-konstanz.de/sub/index.php/sub/article/view/601
% https://people.ict.usc.edu/~gordon/publications/AAAI-SPRING11A.PDF
% https://www.aclweb.org/anthology/P19-1472/
% https://www.aclweb.org/anthology/P16-1144/
% https://www.aclweb.org/anthology/N19-1245/
% https://www.aclweb.org/anthology/N18-1023/
% https://aclweb.org/anthology/D18-1260.pdf
% https://arxiv.org/abs/1911.11641
% https://arxiv.org/abs/1909.06146
% https://aclweb.org/anthology/W17-4413.pdf
% https://arxiv.org/abs/1907.10641

\subsection{Results}

Of the tasks evaluated, several tasks remained near chance or had very high variance, resulting in no clear trend. Of the remainder, an absolute majority exhibited an initial increase in performance and then a decrease in performance after the amount of documents discarded surpassed a threshold that varied by task. Additionally, for almost all tasks the most filtered model was not the best performing. Some tasks like BoolQ exhibit little clear trend. Not all tasks have the same optimal $\alpha$---compare PiQA and LAMBADA---and some tasks like PubmedQA show a much more sudden decrease in accuracy. For results on all tasks, see Figure \ref{fig:montage}.

% \begin{figure}[t]
%     \centering
%     \includegraphics[width=0.5\textwidth]{figures/fig_filterratio_anli_r3.acc.png}
%     \caption{ANLI Round 3 accuracy for various different filtering ratios}
%     \label{fig:fig_filterratio_anli_r3.acc}
% \end{figure}

% \begin{figure}[t]
%     \centering
%     \includegraphics[width=0.5\textwidth]{figures/fig_filterratio_winogrande.acc.png}
%     \caption{Winogrande accuracy for various different filtering ratios. Winogrande is one of the only tasks on which the most aggresively filtered model performed the best.}
%     \label{fig:fig_filterratio_winogrande.acc}
% \end{figure}

\subsection{Analysis}

We hypothesize that this decline in performance is because of misalignment between the classifier objective, intended to be a proxy for quality, and actual document quality. For instance, a classifier to distinguish WebText2 from Common Crawl, as in GPT-3, would also exclude domains of text data not found as often in WebText2. 

We also hypothesize that the difference in optimal $\alpha$ between different tasks is because the characteristics of the different types of data that help the most with each task are over/underdiscarded to a different extent due to spurious correlations with the quality metric. As such, we do not expect the exact thresholds to transfer to other tasks, classifiers, or datasets. This is an expected consequence of Goodharting, because the degree to which different types of text data correlate with the features learned by the classifier is mostly spurious.

\section{Domain Misalignment Experiment}

To test the hypothesis that the misalignment of the objective leads to the exclusion of non-OpenWebText2-like data, we train a fasttext classifier to classify between BookCorpus2 \citep{Gao2020The_Pile__An_800GB_Dataset_of_} and OpenWebText \citep{OpenWebText}, and compute the mean BookCorpus2-probability of each training set. If the classification model is favoring OpenWebText-like data over generally high-quality data, then as filtering increases in intensity, the proportion of BookCorpus2-like data should decrease as the data consists increasingly of OpenWebText-like text. Conversely, if the classification model is robustly favoring high quality text, then as filtering increases in intensity, the proportion of BookCorpus2-like data should \textit{increase}, as low-quality text looks nothing like BookCorpus2. We also repeat this experiment for Pubmed Abstracts. 

We chose BookCorpus2 and Pubmed Abstracts because of their similarity in distribution to LAMBADA and PubmedQA respectively, in the hopes of observing a similarity between the task evaluation curves and the data domain curves.

\subsection{Results}

\begin{figure}[t]
    \centering
    \includegraphics[width=0.5\textwidth]{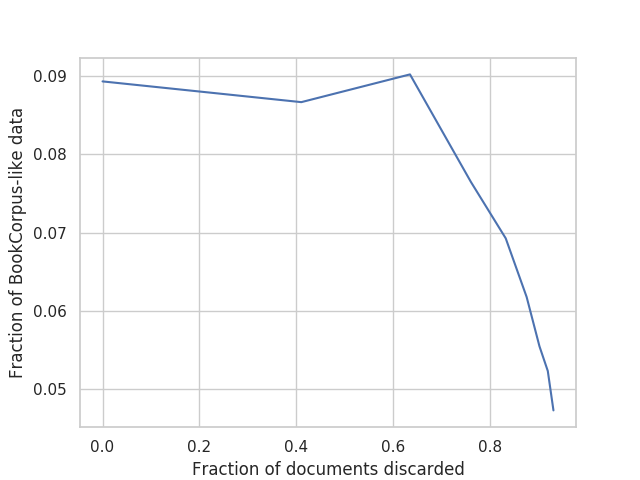}
    \caption{Fraction of documents in filtered Common Crawl classified as BookCorpus2-like by a shallow classifier trained to distinguish OpenWebtext and BookCorpus2. Note that this plot has a different x-axis scale from the task evaluation plots.}
    \label{fig:fig2_bc_sim}
\end{figure}

\begin{figure}[t]
    \centering
    \includegraphics[width=0.5\textwidth]{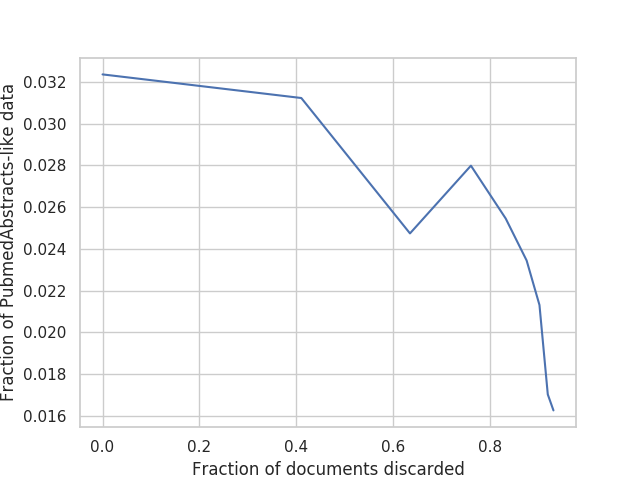}
    \caption{Fraction of documents in filtered Common Crawl classified as PubmedAbstracts-like by a shallow classifier trained to distinguish OpenWebtext and PubmedAbstracts. Note that this plot has a different x-axis scale from the task evaluation plots.}
    \label{fig:fig2_pubmeda_sim}
\end{figure}

As seen in Figure \ref{fig:fig2_bc_sim}, the fraction of BookCorpus2-like data remains mostly constant until around 0.6, after which it declines sharply. A similar pattern is observed with Pubmed Abstracts, albeit with an earlier drop (Figure \ref{fig:fig2_pubmeda_sim}).

The BookCorpus2-like data curve's drop precedes the LAMBADA performance drop by about 0.2. Similarly, the Pubmed Abstracts drop also precedes the PubmedQA's main drop slightly.

\subsection{Analysis}

The decrease in Pubmed Abstracts and BookCorpus2 like data as filtering increases in aggressiveness supports the hypothesis that part of the problem is that text domains not similar to OpenWebText2 are being discarded. 

Our main hypothesis for why the domain data content starts decreasing before the evaluation metric performance does is that these tasks are sufficiently different in distribution to the respective datasets.

%\todo[inline]{Run all the experiments again but this time with a GPT/BERT based classifier}

\section{Limitations}

This work is intended to show that the common assumption that more aggressive data filtering is better is not always true, and thus focuses on one particular classifier used in the real world as an illustrative example. Depending on the type of classifier, the training data used for the classifier, and the downstream task, this effect may not be relevant in certain settings. We leave an exhaustive exploration of the contribution of these various factors to future work.

\section{Conclusion}

In this paper, we explored the effect of filtering the training data using a shallow model trained on a proxy for quality on downstream language model performance. We showed that increasing the aggressiveness of filtering against this signal actually decreases model performance past a certain point, and speculate that this is due to Goodhart's law, as the misalignment between proxy and true objective becomes more significant with increased optimization pressure. We hope that this work leads to more careful analysis of the effects of filtering in future language modeling work.

\section*{Acknowledgements}

The author would like to thank TPU Research Cloud for providing the computational resources for the training, and CoreWeave for providing the computational resources for data processing and evaluation.

The author would also like to thank Stella Biderman, Sid Black, Charles Foster, Eric Hallahan, Kyle McDonell, Jason Phang, and Laria Reynolds for providing feedback on the manuscript.

\bibliography{anthology,custom}
\bibliographystyle{acl_natbib}

% \appendix

% \begin{figure*}
%     \centering
%     \includegraphics[width=\textwidth]{figures/montage.png}
%     \caption{Plots of results for all downstream tasks explored in this paper}
%     \label{fig:montage}
% \end{figure*}

%\section{Example Appendix}
%\label{sec:appendix}

\end{document}